# Improving Diagnostic Accuracy for Oral Cancer with inpainting Synthesis Lesions Generated Using Diffusion Models

**Yong Oh Lee[1], JeeEun Kim[1], and Jung Woo Lee[2,*]**

[1]Department of Industrial and Data Engineering, Hongik University, Seoul, 04066, South Korea
[2]Department of Oral and Maxillofacial Surgery, Kyunghee University, Seoul, 02447, South Korea
*Corresponding: omsace@khu.ac.kr

## ABSTRACT

In oral cancer diagnostics, the limited availability of annotated datasets frequently constrains the performance of diagnostic models, particularly due to the variability and insufficiency of training data. To address these challenges, this study proposed a novel approach to enhance diagnostic accuracy by synthesizing realistic oral cancer lesions using an inpainting technique with a fine-tuned diffusion model. We compiled a comprehensive dataset from multiple sources, featuring a variety of oral cancer images. Our method generated synthetic lesions that exhibit a high degree of visual fidelity to actual lesions, thereby significantly enhancing the performance of diagnostic algorithms. The results show that our classification model achieved a diagnostic accuracy of 0.97 in differentiating between cancerous and non-cancerous tissues, while our detection model accurately identified lesion locations with 0.85 accuracy. This method validates the potential for synthetic image generation in medical diagnostics and paves the way for further research into extending these methods to other types of cancer diagnostics.

## Introduction

Deep learning has become increasingly prevalent in medical imaging analysis, revolutionizing the way that medical images are interpreted and utilized[1]. The integration of deep learning models enables enhanced accuracy and efficiency, tackling complex diagnostic challenges with unprecedented precision.

One of the applications of deep learning in medical image analysis is the generation of synthetic images[2]. Synthetic medical images facilitate the augmentation of scarce datasets, addressing the significant demand for big data in deep learning applications within the medical domain. The scarcity of annotated medical images typically arises from privacy concerns, the rarity of specific conditions, and the substantial time required for expert annotation. These factors lead to limited dataset sizes and considerable variability in pathological features, posing substantial challenges[3]. Medical image datasets often lack the diversity necessary to train robust diagnostic models, resulting in suboptimal performance in practical applications. By generating synthetic lesions, including those in oral cancer images, using advanced deep learning-based generative models, it is possible to enhance both the quantity and quality of the training data, thereby improving the accuracy of diagnostic algorithms[4].

Generative Adversarial Networks (GANs) have been widely utilized in generating synthetic medical images[5]. These models have significantly enhanced dataset expansion, providing ample data for training, and have proven effective in various medical applications. Notably, GANs excel in inpainting medical image modalities, where traditional methods typically fail due to task complexity[6,7]. However, despite their successes, GANs face challenges in handling complex image textures and maintaining details of anatomical or other clinically important features during inpainting tasks. For instance, traditional GANs often struggle with inpainting arbitrarily shaped regions without prior localization, which can lead to inaccuracies in the synthesized medical images[8]. Additionally, maintaining edge and structural integrity in inpainted images remains a significant challenge. These issues can result in overly smoothed textures or distorted structural features, significantly diminishing the clinical utility of the synthesized images[7].

While GANs have significantly advanced synthetic medical imaging, diffusion models[9] emerge as a formidable alternative, particularly suited to the high demands of medical diagnostics. Unlike GANs, which instantaneously generate images, diffusion models incrementally construct images from random noise, meticulously adding detail to achieve high-quality outputs. This methodical process not only enhances stability and image quality but also ensures the preservation of crucial anatomical accuracies, making diffusion models ideal for complex medical imaging tasks[10].

In oral cancer image analysis using deep learning, the challenge has often been the limited availability of annotated datasets, which hampers the development and training of robust diagnostic models. The performance of diagnostic algorithms can be constrained by the quantity and variability of the training data[11,12]. By synthesizing data, researchers can overcome these limitations, providing models with a richer, more varied dataset that mirrors the complexity and diversity of real-world cases. The current methods, while effective in certain scenarios, often fall short in accurately capturing the complex texture and subtle variations in oral lesions[13].

This study introduces a novel application of the stable diffusion model[14] to generate synthetic lesions in oral cancer images using an inpainting approach. Our inpainting technique synthesizes lesions directly within the context of the oral cavity, ensuring that the synthetic images retain realistic textures and anatomical consistency. This method not only enriches the dataset but also enhances the robustness and accuracy of subsequent classification and detection models.

Our study makes several significant contributions to the field of medical image analysis for oral cancer. First, we employ a novel method that uses the stable diffusion model to generate synthetic lesions via inpainting, improving the diversity and quality of data available for training diagnostic models. Importantly, we enhance the diffusion model not by training from scratch but by utilizing fine-tuning techniques, which streamline the adaptation process and ensure higher fidelity in synthetic image generation. Moreover, our approach incorporates segmentation masking with the Segment Anything Model (SAM)[15], which allows for precise definition of lesion areas (masking), overcoming the limitations often faced with datasets that only provide bounding boxes. This method ensures that our synthetic images are not only high-quality but also replicate the detailed structural characteristics of the oral cavity and the specific features of oral cancer lesions. Further, by integrating these synthetic images into existing datasets, we have demonstrated improvements in the accuracy of diagnostic models, verified through rigorous validation against original data. Lastly, our technique's adaptability has been tested across other sources of data, showing consistent enhancements in model accuracy without the need for retraining, thereby proving the scalability and effectiveness of our approach in a variety of settings.

## Methods

### Dataset Description

In this study, we utilized multiple datasets to create a robust foundation for training and testing our deep learning models.

The first dataset, referred as internal dataset, consists of standard photographs of the oral cavity. This dataset, originally named *Final Oral Cancer Merge*, was sourced from the Roboflow data repository[16]. The dataset comprises 1,171 images for train, validation, and test. For this study, only the train data, consisting of 777 images, was utilized. 145 images depict normal oral conditions, while 632 images show the presence of oral cancer lesions. Each image is annotated with bounding boxes to identify the lesions, resulting in a total of 777 annotations across the dataset. All images are fixed to a resolution of 640x640 pixels to ensure consistency across the dataset. Example images are shown in Figure 1.

The second dataset, referred to as the external dataset, combines the oral cavity dataset from Piyarathne et al.[17] with data collected from Kyung Hee University Dental Hospital.**This study was approved by the Institutional Review Board (IRB) of the Kyung Hee University Dental Hospital (approval number: KH-DT23020), and all procedures adhered to relevant ethical guidelines and regulations. Informed consent was obtained from all participants and/or their legal guardians.** This external dataset expands our corpus with an additional 3,000 images, each accompanied by at least one associated mask, totaling 6,358 masks. The images vary in resolution, ranging from 654x597 pixels to 4608x4608 pixels, with 219 images at 4000×1800 resolution and 900 images at 4608×3456 resolution. This diversity reflects variations in image acquisition sources and conditions. For consistency and computational constraints of our models, all synthesized images were generated at a resolution of 512x512 pixels. Example images are shown in Figure 2.

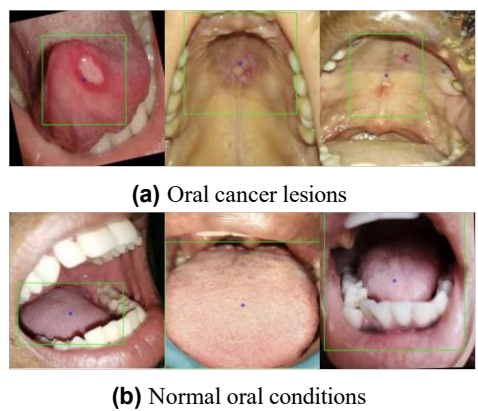

(a) Oral cancer lesions

(b) Normal oral conditions

**Figure 1.** Sample images from the internal dataset depicting oral conditions.

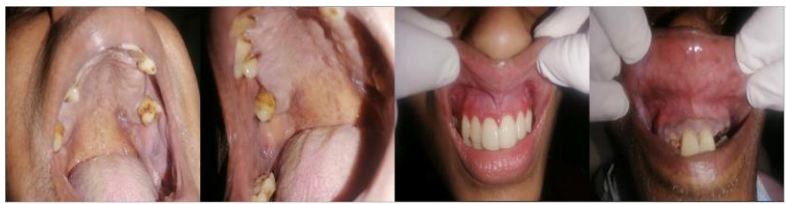

**Figure 2.** Sample images from the external dataset used in the study.

## Masking inpainting Region with Segment Anything Model

To prepare the dataset for our deep learning models, segment masking using the SAM was employed to accurately identify and isolate oral cancer lesions from surrounding tissue. This process was instrumental in enhancing the subsequent stages of model training and image synthesis, particularly for inpainting tasks.

SAM, introduced in its foundational paper[15], represents a significant advancement in the field of image segmentation. As a foundation model designed for diverse applications, SAM has been extensively developed and adapted for various uses, including medical image analysis. Specifically, the SAM ViT-H (Segment Anything Model with Vision Transformer-Huge)[18] version of this tool incorporates around 600 million parameters, providing robust and precise identification of features within images. This makes SAM ViT-H a cutting-edge tool for segmenting complex medical images. Unlike traditional models that often require extensive training and fine-tuning for specific datasets, our implementation of SAM ViT-H did not undergo such fine-tuning. This may affect the precision of lesion masking when compared to expert-annotated masks, a point we explore further in the discussion section.

For the internal dataset, which only contains bounding boxes, the SAM's advanced capabilities were essential for generating accurate lesion masks for inpaining synthesis. The segment masking process began by converting YOLO-formatted bounding box coordinates into pixel values. These pixel values then defined the central point of each lesion, serving as 'point prompts' for the SAM. This approach enabled the SAM to generate detailed masks that closely followed the contours and characteristics of the lesions, effectively isolating them from the non-affected areas of the oral cavity.

In contrast, the external dataset did not require the application of SAM for segmentation. This dataset was originally purposed for segmentation tasks and therefore came pre-masked, eliminating the need for additional processing with SAM. Our use of SAM was exclusively applied to the internal dataset to align its utility with datasets that lacked pre-segmented masks.

The masking facilitated by SAM not only enables more focused analysis by neural networks during training but also enhances the quality of synthetic image generation through the inpainting process. Figures 3 and 4 show examples of masks generated using SAM from the internal dataset, alongside those from the external dataset, to demonstrate the comparative effectiveness and application of SAM in different dataset contexts.

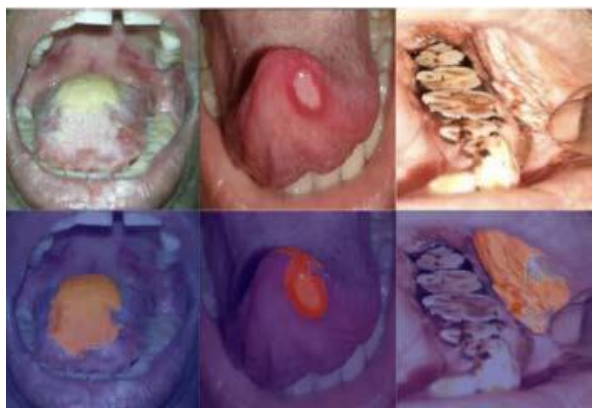

**Figure 3.** Example of original oral images and their corresponding lesion masks generated using SAM from the internal dataset. The top row presents the original images showcasing various oral cancer lesions. The bottom row illustrates these images after the application of segmentation masks (highlighted in red).

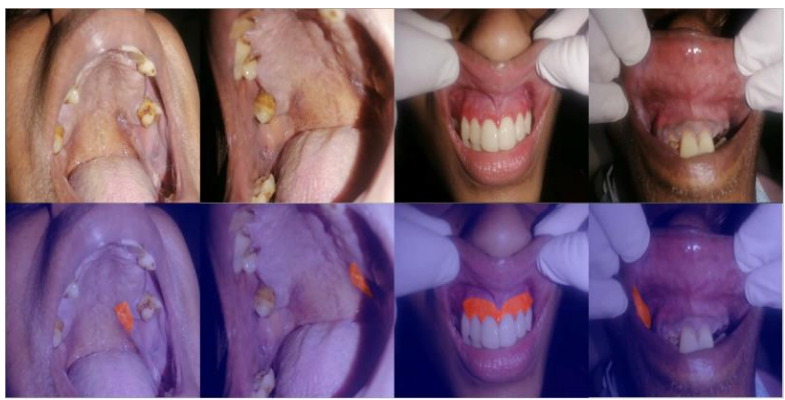

**Figure 4.** Example of a pre-segmented mask from the external dataset. The top row presents the original images showcasing various oral cancer lesions. The bottom row illustrates their segmentation masks (highlighted in red). This illustrates the initial segmentation provided, which did not require further processing with SAM, highlighting differences in dataset preparation.

**Fine-tuning Diffusion Model**

For the fine-tuning of stable diffusion model, we leveraged the DreamBooth[19] to tailor the model specifically for oral cancer lesion imagery. This customization is essential for generating synthetic images that are not only high in quality but also detailed and specific to the features typical of oral cancer. The fine-tuning process followed several key steps, detailed as follows.

**Instance Data Preparation**: We used a subset of our internal dataset, specifically the training portion of the final-oral-cancer-merge, which consists of 777 images. These images provide a comprehensive representation of oral cancer lesions and are crucial for informing the model about the specific attributes of the target lesions.

**Prompt Specification**: Each image in the training set was associated with the prompt 'a medical image of an oral cancer lesion'. This prompt helps to focus the model's learning process on the specific task of generating medical-grade images that accurately reflect the complex characteristics of oral cancer lesions.

We fine-tuned the stable diffusion model with meticulous attention to detail, crucial in medical imaging applications. Each image from the training dataset was processed individually with a batch size of one. This approach ensured intense focus on each image, allowing for precise adjustments and deep learning from the unique characteristics of each lesion depicted. The learning rate was set to 5e-6 to facilitate gradual and accurate learning, capturing the nuanced details of the medical images without overshooting, which is vital for maintaining the integrity and diagnostic utility of the generated images. The model underwent 1,000 training steps, strategically chosen to prevent overfitting while adequately adapting to our specific needs in oral cancer imaging.

This fine-tuning imbues the stable diffusion model with the capability to generate detailed and anatomically precise representations of oral cancer. These capabilities are essential not only for improving diagnostic models but also for providing training data where actual medical images are scarce. Through fine-tuning, the model learns to interpret and synthesize complex medical images, necessary for advancing oral cancer diagnostics.

**Diffusion inpainting for Synthetic Oral Cancer Lesion Generation**

We applied diffusion inpainting with original oral cavity images combined with lesion masks generated by SAM. This method precisely targets lesions and preserves the surrounding areas during the inpainting process. For our internal dataset, which lacks pre-segmented masks, SAM identifies and points out the central points of lesions. This capability ensures accurate lesion detail replication and correct alignment of inpainting with the actual lesion locations.

We directed the stable diffusion model, which is fine-tuned with Dreambooth, using a positive prompt, 'a medical image of oral squamous cell carcinoma' to generate images that closely mimic the appearance and pathological features characteristic of oral cancer. To ensure the quality of the images, a negative prompt 'blurry, low quality' was used, which helped in avoiding undesired attributes such as blurriness and low resolution in the synthetic images. The final images were processed at a resolution of 512x512 pixels, a choice that offers a good balance between detail and computational efficiency, producing three variations per input to provide a comprehensive view of the synthetic lesions. The inference process was fine-tuned with 100 steps, allowing the model to elaborate on details and textures critical for a realistic representation of the disease. A guidance scale, which significantly influenced the model's output towards generating clinically relevant features while minimizing irrelevant variations, is set to 7.5. Figure 5 illustrates the diffusion inpainting procedure used to generate synthetic lesions.

Figure 6 demonstrated that the proposed method replicates complex textures and colors associated with oral lesions, enhancing the potential for these images to be used in oversampling for the diagnosis model training. Figure 6a shows outcomes derived from our internal dataset, wherein the diffusion model underwent fine-tuning. The generated images preserve the distinctive visual attributes of oral cancer lesions while integrating subtle modifications into the synthesized imagery. This approach guarantees that the generated images, despite their diversity, remain faithful to the essential textures and color patterns critical for precise model training. Figure 6b presents the synthesis results utilizing an external dataset, which did not require segment masking by SAM. Although this dataset was not employed in fine-tuning the diffusion model, it exemplifies the model's capacity to effectively render lesions. The diffusion inpainting model's adeptness in replicating intricate lesion textures, underscores its robustness and demonstrates its applicability to other oral cavity images.

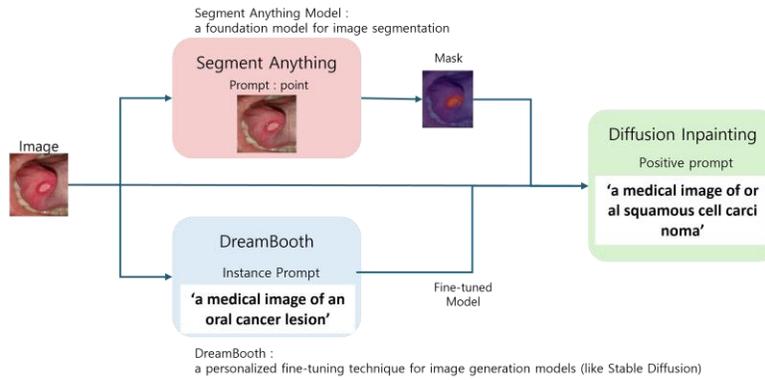

**Figure 5.** Overview of the diffusion inpainting process for synthetic oral cancer lesion generation. The workflow starts with original oral cavity images, which are segmented by SAM to delineate lesions. These regions are then processed by a fine-tuned diffusion model to synthesize lesions.

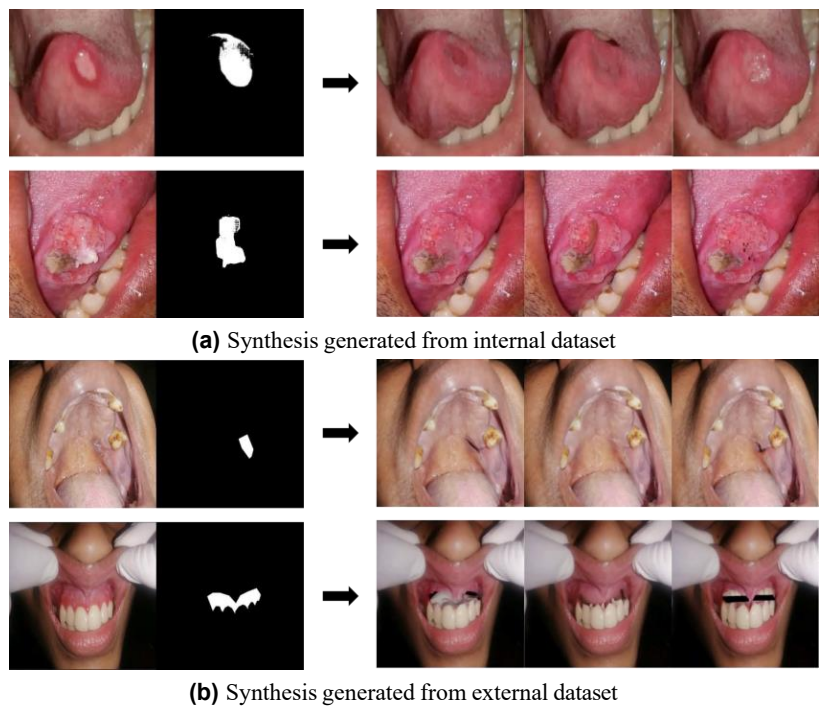

(a) Synthesis generated from internal dataset

(b) Synthesis generated from external dataset

**Figure 6.** Diffusion-inpainting results using the external dataset. The left panels display the original images and their corresponding masks, while right panels showcase the synthetic images produced. This visualization highlights the model's capacity to replicate detailed features critical for clinical analysis and diagnosis in oral cancer.

# Results

## Analysis of Generated Image Quality

We conducted a quantitative analysis of the generated images quality using both internal and external datasets to assess their quality through various established metrics. These metrics include the Peak Signal-to-Noise Ratio (PSNR), the Structural Similarity Index (SSIM), the Learned Perceptual Image Patch Similarity (LPIPS), and the Fréchet Inception Distance (FID)[14]. Each of these metrics provides insights into different aspects of image quality from numerical accuracy to perceptual similarity.

The PSNR, which measures the peak signal-to-noise ratio in decibels, showed values of 24.13 dB for the internal dataset and 21.23 dB for the external dataset. These scores suggest a reasonable level of accuracy in image reproduction, with the internal dataset exhibiting slightly better performance. The SSIM values, which assess structural similarity, were 0.8503 for the internal dataset and 0.8228 for the external, indicating a high degree of structural accuracy in the synthetic images, especially from the internal dataset. When focusing on the inpainting areas, the PSNR values increase slightly to 28.34 dB for the internal dataset and 28.24 dB for the external dataset, reflecting a modest enhancement in signal accuracy. However, the SSIM values in these regions show a slight decrease, registering at 0.7420 for the internal and 0.7192 for the external dataset, indicating a mild reduction in structural fidelity. For the more perceptually oriented LPIPS, which uses deep learning features to evaluate image similarity, the scores were 0.1378 for the internal dataset and 0.2396 for the external. Lower scores in LPIPS denote closer perceptual resemblance to the original images, highlighting the internal dataset's superior fidelity in maintaining perceptual qualities. Lastly, the FID scores, which compare the distribution of features extracted by an Inception network, were 32.74 for the internal dataset and 21.51 for the external dataset. Lower FID scores indicate a smaller distance between the distributions of generated and original images, suggesting that both datasets achieved a commendable level of consistency, with the external dataset performing notably well.

These results collectively affirm that the generated images not only closely mimic the original datasets in terms of structural and perceptual quality but also demonstrate the model's effectiveness across different data sources. This is crucial for medical imaging applications where the fidelity and reliability of synthetic images are paramount for accurate diagnosis and treatment planning. The superior metrics from the internal dataset particularly underscore the model's capability to handle detailed and varied medical image features effectively, making it a promising tool for enhancing diagnostic imaging practices.

**Table 1.** Quantitative Analysis Results of Generated Images

| Dataset | PSNR (dB) | SSIM | LPIPS | FID |
|---|---|---|---|---|
| Internal Dataset | 24.13 | 0.8503 | 0.1378 | 32.74 |
| External Dataset | 21.23 | 0.8228 | 0.2396 | 21.51 |

## Enhancement of Classification with Generated Synthesis

The inclusion of synthetic images significantly enhanced the classification, as evidenced by the results obtained from a 5-fold cross-validation analysis using a ResNet-50 model[20] trained on our internal dataset. In this setup, only images depicting oral cancer conditions were over-sampled, producing four times the number of synthetic images compared to the original, which effectively increased the dataset from 777 to a much larger pool focused on enhancing model exposure to varied cancerous conditions.

When trained solely with the original dataset, the model's Area Under the Receiver Operating Characteristic curve (AUROC) ranged from 0.8907 to 0.9825 with an overall average of 0.9508, reflecting considerable variability. The inclusion of synthetic images in the training improved the AUROC slightly, with values ranging from 0.9447 to 0.9722 and an overall average increasing to 0.9586, indicating both an increase in performance and a reduction in variability.

Similarly, the accuracy when training with only the original dataset varied from 0.9065 to 0.9714, with an average of 0.9228, showing significant fluctuation. By incorporating synthetic data, accuracy increased to a range between 0.9584 and 0.9792, with the average climbing to 0.9705. This shift not only raised the mean accuracy but also noticeably decreased its variation across folds, highlighting the effectiveness of using synthetic data to enhance classification model consistency and reliability.

Table 2 encapsulates these findings, illustrating the enhancement in model performance due to the addition of synthetic data.

As illustrated in Figure 7, the Grad-CAM[21] images provide a clearer visualization of the impact of using synthetic data. When training solely with the original data, the areas activated in the Grad-CAM results sometimes extend beyond the segment masking boundaries. In contrast, with training that includes oversampling using synthetic data, the activated regions are confined well within the boundaries of the segment masking. This demonstrates that the synthetic data helps focus the model's attention more accurately on relevant lesion areas, enhancing the precision of the diagnostic model.

**Table 2.** Comparison of Classification Metrics With and Without Synthetic Data in model training

| Test Metric | Train Data | Fold-1 | Fold-2 | Fold-3 | Fold-4 | Fold-5 | Overall(mean ± sd.) |
|---|---|---|---|---|---|---|---|
| AUROC | Original Images | 0.9676 | 0.9825 | 0.9676 | 0.8907 | 0.9459 | **0.9508 ± 0.0323** |
|  | Origin + Synthetic Images | 0.9447 | 0.9559 | 0.9538 | 0.9664 | 0.9722 | **0.9586 ± 0.0097** |
| Accuracy | Original Images | 0.9714 | 0.9643 | 0.9143 | 0.9143 | 0.9065 | **0.9228 ± 0.0277** |
|  | Origin + Synthetic Images | 0.9709 | 0.9584 | 0.9792 | 0.9792 | 0.9647 | **0.9705 ± 0.0081** |

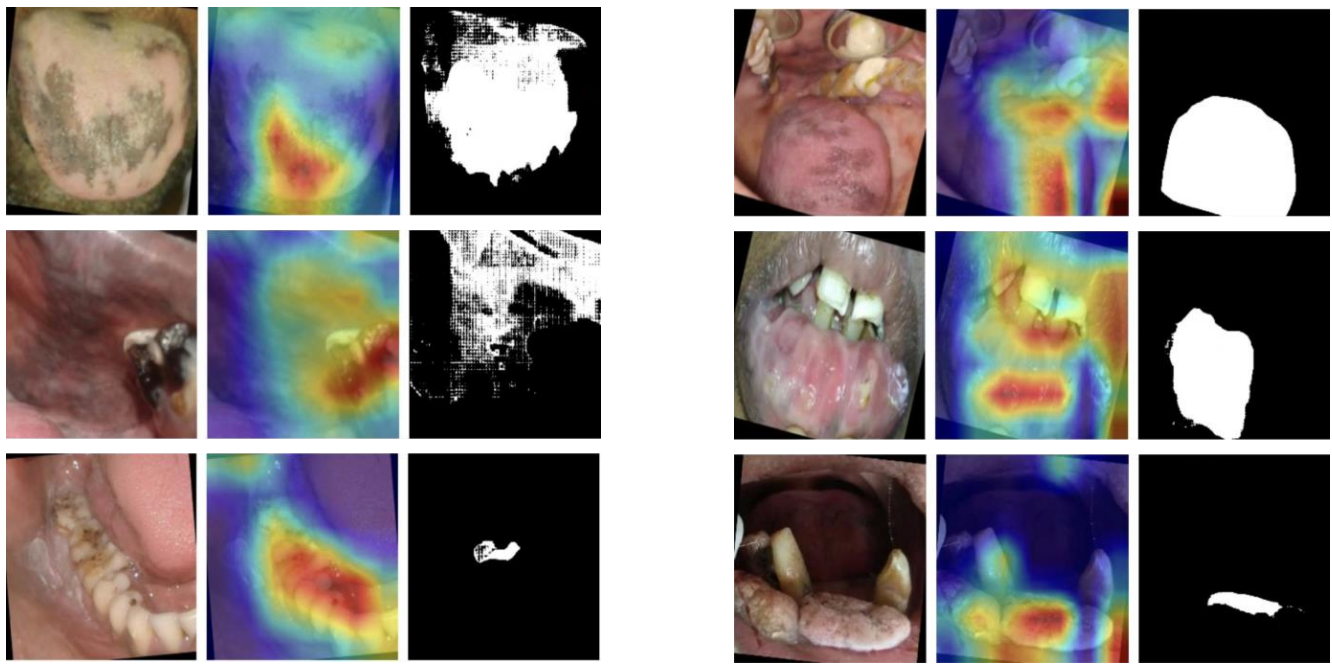

**Figure 7.** Grad-CAM results for various oral lesions: On each side, the first image is the original image, the second shows the Grad-CAM heatmap, and the third displays the segment masking area created by SAM. The left side illustrates results from training with only the original data, while the right side shows results from training with both original data and synthetic data.

**Enhancement of Detection with Generated Synthesis**

In this section, we examine the effects of synthetic data oversampling on the detection of oral cancer lesions using the YOLO v8 model[22]. The model underwent training for 120 epochs, configured with a batch size of 4 and an image resolution of 640 pixels. We set the learning rate at 0.005 and utilized a momentum of 0.95, optimizing the process with Stochastic Gradient Descent (SGD). The results of comparison of detection model's performance between training with or without synthetic data are shown in Table 3.

When synthetic images were incorporated into the dataset, precision increased from 0.788 to 0.851. This improvement indicates that the models became more accurate in identifying true positives, an enhancement likely attributed to the additional variety and complexity provided by the synthetic images. This result is particularly valuable in clinical settings where the cost of false positives is high, suggesting that synthetic augmentation can lead to more reliable diagnostic predictions.

On the other hand, recall experienced a marginal decrease from 0.783 to 0.767. This slight reduction suggests that while the models became more precise, they might miss a few positive cases they previously captured. However, this small decrease is compensated by significant gains in precision and overall accuracy, which are crucial for clinical applications where precision is more critical than recall.

The metrics for mAP at 50 IoU (mAP 50) and mAP from 50 to 95 IoU (mAP 50-95) both showed improvements. The mAP at 50 IoU increased from 0.840 to 0.869, and the mAP at 50-95 IoU rose from 0.566 to 0.587. These enhancements confirm that the model's ability to correctly localize and detect objects has improved, particularly at higher thresholds of intersection over union, which is essential for precise medical diagnostics.

The integration of synthetic images into the training sets has proven to be highly effective in enhancing the detection capabilities of our models. This strategy not only improves precision but also enhances the model's ability to accurately localize lesions within a varied range of scenarios, thereby increasing the reliability and utility of automated diagnostic tools in medical imaging. These findings advocate for the continued use and exploration of synthetic data augmentation in improving the performance of detection models, particularly in domains where accuracy and precision are paramount.

**Table 3.** Comparison of Detection Metrics With and Without Synthetic Data

| Metric | Original Images | Original + Synthetic Images |
|---|---|---|
| Precision | 0.788 | 0.851 |
| Recall | 0.783 | 0.767 |
| mAP 50 | 0.840 | 0.869 |
| mAP 50-95 | 0.566 | 0.587 |

In Figure 8, we demonstrate the YOLO v8 model's ability to detect oral cancer lesions with varying confidence levels, ranging from 0.6 to 0.9. These images, from synthetically augmented datasets, illustrate the model's enhanced precision in identifying cancerous lesions.

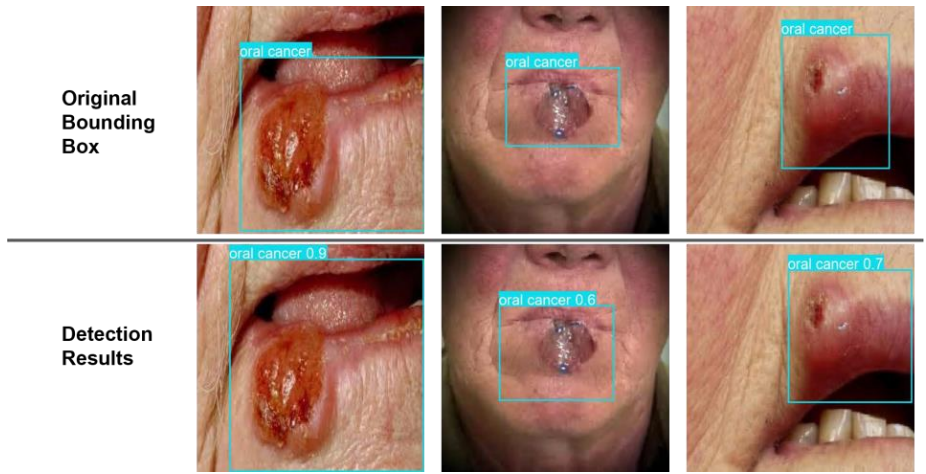

**Figure 8.** Detection of oral cancer lesions across a variety of cases. The top row shows original bounding box as annotations, and the bottom row includes results YOLO v8's predicted bounding box when the model trained with oversample data using synthesis, illustrating the model's accuracy and confidence in detection across different cases.

## Conclusion

This study has demonstrated that incorporating synthetic images generated by stable diffusion models enhances the performance of diagnostic models for oral cancer. The use of synthetic oversampling in training has led to improvements in both classification and detection metrics, including increases in AUROC, accuracy, and mean mAP. These enhancements confirm the utility of diffusion-based image synthesis in enriching training datasets for medical AI, which often face limitations due to the scarcity of diverse medical images.

The integration of synthetic images not only improves the accuracy and reliability of diagnostic models but also suggests a broader applicability of this method across various domains of medical imaging. The success of this approach highlights the potential for synthetic data to compensate for the deficits in existing datasets, thereby enabling the development of more robust models. For instance, the combined use of advanced medical segmentation model[23] for inpainting region extraction and the fine-tuning of diffusion models[24] underscores the practical effectiveness of synthetic data in advancing medical diagnostics.

Looking ahead, there is potential for extending the application of synthetic data generation to other areas of medical imaging. Future research will explore the applicability, benefits, and limitations of these techniques in different medical contexts. As the SAM model evolves and becomes optimized for medical data, its utility is expected to increase, particularly when applied to foundational models for medical learning beyond specific disease diagnostics. This progression promises to accelerate disease detection, enhance treatment efficiency, and ultimately improve patient outcomes.

## Acknowledgements

This research was supported by a grant of the Korea Health Technology R&D Project through the Korea Health Industry Development Institute (KHIDI), funded by the Ministry of Health & Welfare, Republic of Korea (grant number: HI23C0162). **For dataset from Kyung Hee University Dental Hospital, the Institutional Review Board (IRB) of the Kyung Hee University Dental Hospital provided approval (KH-DT23020).**

## Author contributions statement

Y.O. Lee and J.W.Lee conceived the experiment(s), Y.O.Lee and J.Kim. conducted the experiment(s), Y.O.Lee, J.Kim and J.W.Lee analysed the results. All authors reviewed the manuscript.

## Additional information

**Requests for the data supporting the findings of this study should be directed to the corresponding author. The data may be made available upon reasonable request and following review and approval.**